\documentclass[runningheads]{llncs}
\usepackage[utf8]{inputenc}
\usepackage{xspace}
\usepackage{multirow}
\usepackage{todonotes}
\usepackage[export]{adjustbox}
\usepackage[final,pdfpagelabels=false,pdfpagelayout=SinglePage,bookmarksnumbered,hyperindex,colorlinks,linkcolor=blue,citecolor=blue,urlcolor=blue]{hyperref}
\usepackage[nameinlink]{cleveref}
\usepackage{subcaption}
\usepackage{listings}

\usepackage{inconsolata} %

\lstdefinestyle{essence}{
 basicstyle=\ttfamily\scriptsize 
, keywords = {language, \essence, given, letting, find, such, that, where
, domain, function, total, surjective, be
, forAll, exists, injective, in, preImage, range
, int, mset, set, partition, new, type, intersect, matrix, from
, minimising, maximising, indexed, by, bool
, defined, minSize, maxSize, size, maxNumParts, numParts
, subset, subsetEq, relation
, toInt, sum
}
, frame = single
, framesep = 5pt
, morecomment=[l]{\$} 
}

\lstdefinestyle{PDDL}{
 basicstyle=\scriptsize 
, keywords = { predicates, action, parameters, precondition, effect
, init, domain, objects, define, and, goal, problem, instance-constraints
, forall, exactly-k, atleast-k, appear, not
}
, frame = single
, framesep = 5pt
, morecomment=[l]{\;} 
}

\newcommand{\comment}[1]{} \newcommand{\code}[1]{{\small\texttt{#1}}} \newcommand{\conjure}{\textsc{Conjure}\xspace} 
\newcommand{\essence}{\textsc{essence}\xspace} 
\newcommand{\savilerow}{Savile Row\xspace}

\newcommand{\minion}{\code{minion}\xspace} 
\newcommand{\irace}{\textsf{irace}\xspace} 
 
\newcommand{\rantanplan}{rantanplan\xspace} 
\newcommand{\floor}{\texttt{floor-tile}\xspace}

\newcommand{\name}[1]{\texttt{#1}}

\title{Exploring Instance Generation\texorpdfstring{\\}{ }for Automated Planning}
\titlerunning{Exploring Instance Generation for Automated Planning}

 \author{\"Ozg\"ur Akg\"un \and Nguyen Dang \and Joan Espasa \and Ian Miguel \and Andr\'as Z. Salamon \and Christopher Stone}
 \authorrunning{Akg\"un et al.}

 \institute{
 School of Computer Science, University of St Andrews, UK\\
 \email{\{ozgur.akgun,nttd,jea20,ijm,Andras.Salamon,cls29\}@st-andrews.ac.uk}
 }

\date{\now{}}

\begin{document}

\maketitle

\begin{abstract}
Many of the core disciplines of artificial intelligence have sets of standard benchmark problems well known and widely used by the community when developing new algorithms. Constraint programming and
 automated planning are examples of these areas, where the behaviour of a new algorithm 
 is measured by how it performs on these instances. Typically the efficiency of each solving method
  varies not only between problems, but also between instances of the same problem. 
 Therefore, having a diverse set of instances is crucial to be able to effectively evaluate a new 
 solving method. Current methods for automatic generation of instances for Constraint Programming problems start with a declarative model and search for instances with some desired attributes, such as hardness or size. 
We first explore the difficulties of adapting this approach to generate instances starting from problem specifications written in PDDL, the de-facto standard language of the automated planning community.
We then propose a new approach where the whole planning problem description is modelled using \essence, an abstract modelling language that allows expressing high-level structures without committing to a particular low level representation in PDDL.
\end{abstract}

\section{Introduction}

The planning task consists of selecting a sequence of actions in order to achieve a specified goal from 
specified starting conditions. This type of problem arises \comment{frequently }in many contexts. Consider, for
example, the delivery of a set of packages by vehicle from a depot to a set of destinations.
The allocation of packages and drivers to vehicles must be planned, as well as the
route for each vehicle, while respecting package delivery deadlines, vehicle capacities and driver shift
restrictions.

Given their importance, the automated solution of planning problems is a central discipline of Artificial
Intelligence. The difficulty of solving planning problems grows rapidly with their size in terms of the 
number of objects and possible actions under consideration. Over many years, a great deal of effort by  different research groups has resulted in the development of highly efficient AI planning 
systems~\cite{vallati20152014}.
Testing algorithms across a wide range of problem instances is crucial to ensure the validity of any claim 
about one algorithm being better than another. However, when it comes to evaluations, typically limited 
sets of problems are used and thus the full picture is rarely seen. Finding and encoding interesting instances 
is a time-consuming task, and due to the nature of some problems, is sometimes out of reach from the researcher
perspective.

In the International Planning Competitions (IPC) state of 
the art planning systems are empirically evaluated on a set of benchmark problems. The competitions have,
amongst others, a track that focuses on planners that can learn from previous runs. This track uses manually coded problem generators to provide the planners with a varied set of problems. This variety of problems is crucial to ensure that planners can learn and generalise well on new and unseen situations.

The problem of automated instance generation is to make the process of creating benchmark problems more accessible and efficient.
Instance generation has been a focus of the SAT community for several decades~\cite{SelmanML96}. Generating random SAT instances while considering different parameters such as the length of clauses, the number of variables, or their connectivity, has helped to illuminate how algorithms behave and how they perform in different circumstances.
Having automated tools to generate interesting instances allows algorithm developers to evaluate and compare the algorithms across a wide range of instances, providing a detailed picture of their comparative strengths and weaknesses.

In this work we explore the adaptation of a successful tool~\cite{Akgun2019:instance} that automatically generates instances with desirable properties from a single problem specification in \essence, a high-level modelling language for Constraint Programming (CP)~\cite{Akgun2011:extensible}.
We discuss two approaches (Section~\ref{sec:augmented_PDDL} and \ref{sec:instance-gen-in-essence}) to make the instance generation process in~\cite{Akgun2019:instance} work with PDDL~\cite{pddl}, the de-facto standard language in the automated planning community. Both approaches have their own limitations regarding \emph{flexibility}, \emph{efficiency} and \emph{automation}. We then make a proposal for a new planning modelling language using \essence (Section~\ref{sec:essence-for-planning}). Thanks to the rich expressiveness of the language, all discussed limitations of the PDDL-based instance generation approaches are overcome.

\section{Related Work}
\label{sec:relatedwork}

Fuentetaja et al.~\cite{fuentetaja2012planning} approach the generation of satisfiable planning instances as a planning 
problem, where users manually write some declarative semantics-related information to describe the generation of different instances. 
They have pointed out the limitation of PDDL as a representation for describing the instance generation problem, as the task of generating valid initial states is complex and requires information about the meaning of predicates not expressed in the domain definition. PDDL representation is only effective for representing valid states and the transition functions between them. 
Augmenting PDDL to improve its expressiveness has been proposed several times~\cite{KautzS98,BacchusK00,haslum2003domain}. This allows adding extra information into the domain definition, and could potentially lead to more general applications of planning~\cite{positionpddl}.

The usefulness of automated generation of benchmark instances has been demonstrated in various fields, such as combinatorial optimisation~\cite{smith2012measuring}, SAT~\cite{horie1997hard} and model fitting~\cite{munoz2018instance} especially in the context of instance space analysis~\cite{smith2014towards}. In CP, benchmark instances created using problem-specific generators have been proposed for various problem classes, such as binary Constraint Satisfaction Problem~\cite{van2003evolving,moreno2012challenging} and Random Constraint Networks~\cite{xu2007random}. In Operations Research, several instance generators and benchmarks for well-known combinatorial optimisation problems, such as the Knapsack problem~\cite{julstrom2009evolving} and the Nurse Rostering problem~\cite{vanhoucke2009characterization}, have been provided and widely used. All of those approaches are semi-automated, as the instance generators were manually created.

In discrete optimisation Ullrich ~\cite{ullrich2018generic} developed a tool for the generation of instances in the TSP, Max-SAT and Load Allocation problems. Even in this case generating instances for new problem classes requires some user input, however the formal language they developed simplifies the process of producing instances for new domains.

\begin{figure*}[!h]
    \includegraphics[width=\linewidth]{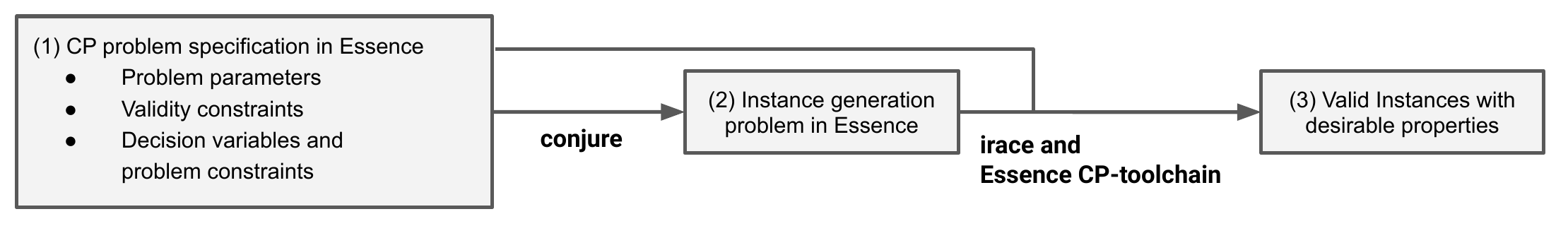}
    \caption{The automated instance generation approach for CP problems proposed in~\cite{Akgun2019:instance,Akgun2020:discriminating}}
    \label{fig:cp-instGen}
    \centering
\end{figure*}

Recently, a new approach for fully automated instance generation has been proposed for CP~\cite{Akgun2019:instance,Akgun2020:discriminating}. The approach allows users to declaratively describe a CP problem and properties of valid instances in a single CP model using the high-level constraint modelling language \essence~\cite{Akgun2011:extensible}. 
An instance generator is then created by the automated modelling tool \conjure~\cite{Akgun2013:automated}. Finally, instances with desirable properties are generated through a combination of the automated algorithm configuration tool \irace~\cite{Lopez-Ibanez2016:irace} and the \essence constraint modelling toolchain developed by the CP group at St Andrews~\footnote{\url{https://constraintmodelling.org/}}, 
which includes \conjure, \savilerow~\cite{Nightingale2017:automatically} (a constraint modelling assistant) and \minion~\cite{Gent2006:minion} (a CP solver). All steps are done in a completely automated fashion. The whole process of the system is shown in Figure~\ref{fig:cp-instGen}.

\section{Background}

A classical planning problem is defined as a tuple $\Pi = \langle V,A,I,G\rangle$ where $V$ is a set of 
propositions (or Boolean variables), $A$ is a set of actions, $I$ is the initial state and $G$ is a formula over $V$ that any goal state must satisfy.

A state is a total assignment to the variables. Actions are formalized as pairs $\langle p,e\rangle$,
where $p$ is a set of preconditions and $e$ a set of effects. More formally, $p$ is a set of Boolean expressions over
$V$, while $e$ is a set of assignments. An action $a = \langle p,e\rangle$ is executable in a given state
$s$ if $s \models p$.%
The state resulting from executing action $a$ on state $s$ is denoted by $apply(a,s) = s'$.
The new state $s'$ is defined by assigning new values to the variables according to the active effects, and
retaining the values of the variables that are not assigned values by any of the active effects.
A plan of length $n$ for a planning problem $\Pi$ is a sequence of actions $a_1;a_2;\dots;a_n$ such that
$apply(a_n,\dots,apply(a_2,apply(a_1,I))\dots) \models G$.

An example planning problem is the \floor problem used in the International Planning Competition 2014 (IPC-14)~\cite{vallati20152014}. The problem includes a set of robots sharing the task of painting a pattern on floor tiles. The robots move around in four directions (up, down, left and right) and can paint with one colour at a time.
They are also able to change the current colour to any other one. However, 
due to their design robots can only paint the tiles that are in front or behind them.
Finally, once a tile has been painted, no robot can stand on it.

Automated planning models are typically expressed in the Planning Domain Definition Language
(PDDL)~\cite{pddl}. The user describes the problem in a \emph{domain} file, in terms of predicates and actions with parameters. 
In turn, these parameters (or free variables) can be instantiated with a set of defined objects.
A concrete instance is expressed in a \emph{problem} file, where the initial state, problem
objects and goal are defined. Figure~\ref{fig:pddl_domain} shows how predicates and actions 
are declared in a domain file for the \floor problem (a problem description model), and Figure~\ref{fig:pddl_instance} depicts an example problem file (an instance).

\begin{figure}[!h]
\centering
\begin{lstlisting}[style=PDDL,breaklines]
(:predicates ; state variables are defined in the predicates section
        (robot-at ?r - robot ?x - tile) ; at what tile a robot is
        (up ?x - tile ?y - tile)
        (down ?x - tile ?y - tile)
        (right ?x - tile ?y - tile)
        (left ?x - tile ?y - tile)
        (clear ?x - tile) 
        (painted ?x - tile ?c - color) 
        (robot-has ?r - robot ?c - color)
        (available-color ?c - color)) 

(:action move_up 
  :parameters (?r - robot ?from - tile ?to - tile)
  :precondition (and (robot-at ?r ?from) (up ?to ?from) (clear ?to))
  :effect (and (robot-at ?r ?to) (not (robot-at ?r ?from))
               (clear ?from) (not (clear ?to))))
...
\end{lstlisting}
\caption{Snippets of the \floor PDDL problem description.}
\label{fig:pddl_domain}
\end{figure}

\begin{figure}[!h]
\centering
\begin{lstlisting}[style=PDDL,breaklines]
(define (problem toy)
 (:domain floor-tile)
 (:objects tile_0-0 tile_0-1  
           tile_1-0 tile_1-1 - tile
           robot1 robot2 - robot
           white black - color)
 (:init 
   (robot-at robot1 tile_0-1) (robot-has robot1 white)
   (robot-at robot2 tile_1-1) (robot-has robot2 black)
   (available-color white) (available-color black)
   (clear tile_0-0) (clear tile_1-0)
   (up tile_0-1 tile_1-1) (up tile_0-0 tile_1-0)
   (down tile_1-1 tile_0-1) (down tile_1-0 tile_0-0)
   (right tile_0-1 tile_0-0) (right tile_1-1 tile_1-0)
   (left tile_0-0 tile_0-1) (left tile_1-0 tile_1-1)
)
 (:goal (and (painted tile_0-0 white) (painted tile_1-0 black))))

\end{lstlisting}
\caption{An example \floor instance.}
\label{fig:pddl_instance}
\end{figure}

\section{Validity Constraints for Planning Instances} %
\label{sec:valid_planning_problems}

A planning model is typically an abstraction of a real-world problem, where the modeller has some
predefined assumptions on how the real-world works. The correctness of this abstraction derives
from a valid initial state, and that the transitions of the state variables between steps respect
the implicit constraints. When a model and an instance respect all the implicit assumptions by 
the modeller we say that it is a \emph{valid} problem.
As an example, some implicit assumptions by the modeller in the \name{floor-tile} domain 
(Figure~\ref{fig:pddl_domain}) are that in the initial state each robot must be at exactly one tile,
or that any given cell can only have one unique cell on top of it. Moreover, in the IPC-14 published instances, the tiles' structure 
(represented by \code{up}, \code{down}, \code{right} and \code{left}) always forms a square grid. 

In the automated planning community, sometimes when a problem is released, a Python or Java program is 
included to generate instances automatically. 
The validity properties of the problem are normally hard-coded in those programs, and therefore appear implicitly in the generated problem instances.

An alternative approach is to allow modellers to express the validity properties of a planning problem declaratively. An instance generator with those validity constraints integrated is then 
automatically created. Specifying those properties using a declarative modelling language provides flexibility, 
as users can easily add or update the validity specification without having to modify the generator software's 
source code directly. This is the approach of the \essence-based automated instance generation system that we propose to build upon.

There is a large variety of validity properties arising in classical planning domains. In this section, we
discuss six arbitrarily selected IPC-14 problems (in the Sequential track) and the validity properties of their
published benchmark instances. A brief description of these problems follows. We refer to the competition
website\footnote{\url{https://helios.hud.ac.uk/scommv/IPC-14/}} and the accompanying paper~\cite{vallati20152014}
for further details of these benchmarks.

\begin{itemize}
    \item \texttt{city-car}: This problem simulates the impact of road building and demolition on traffic
    flows. A city is represented as a grid, in which each node is a junction and edges are potential roads.
    Cars start from different positions and have to reach their final destinations. A finite number of roads can be
    built to connect two junctions and allowing a car to move between them. 
    \item \texttt{floor-tile}: A set of robots use colours to paint patterns in floor tiles. The robots can move
    around and paint with one colour at a time and can also change their colours. Once a tile has been painted, no robot can stand on it. 
    \item \texttt{hiking}: This problem simulates a walking trip that lasts several days, where each day one walk 
    is done with a partner. The walks are over a long circular route, without ever walking backwards. Tents and
    items of luggage can be carried in a car between the start and end of the walking routes if necessary.
    \item \texttt{cave-diving}: There are divers with different skills and confidence, and each can
    carry tanks of air. These divers must be hired to go into an underwater cave system and either take photos or
    prepare the way for other divers by dropping full tanks of air. The cave is too narrow for more than one diver
    to enter at a time. Divers must exit the cave and decompress at the end. They can therefore only make a single
    trip into the cave. Divers have hiring costs inversely proportional to how hard they are to work with.
    \item \texttt{child-snack}: This involves making and serving sandwiches with various ingredients
    for a group of children in which some are allergic to gluten. 
    \item \texttt{barman}: A robot barman manipulates drink dispensers, glasses, and a shaker. The goal is to find
    a plan of the robot's actions that serves a desired set of drinks. Robot hands can only grasp one object at a
    time and glasses need to be empty and clean to be filled. 
\end{itemize}

The properties in these benchmark instances can be roughly divided into two groups. The first group involves
constraints between the variables emerging from a single predicate. They typically include three types of
constraints: \emph{exactly-k}, \emph{at-most-k} and \emph{at-least-k}, where $k$ is a parameter of the constraints. For example, from the \floor model an instances illustrated in Figure~\ref{fig:pddl_domain} and \ref{fig:pddl_instance}, we can infer that each robot begins at 
exactly one tile. This could be represented by, for a given robot, an \emph{exactly-1} constraint
on the variables that result from grounding the \texttt{robot-at} predicate.

The second group of validity constraints involves \emph{structural constraints} between predicates in a planning
problem. The published instances of both \texttt{city-car} and \texttt{floor-tile} have \emph{square-grid 
underlying maps}. The maps of \texttt{floor-tile} require rectangular grids with up, down, left and right connections,
while the maps in \texttt{city-car} instances allow cells to be connected horizontally as well as diagonally.
In \texttt{cave-diving} a cave forms a tree-shaped structure.
Sequences are represented in the \texttt{hiking} and \texttt{barman} domains, while other problems from the competition use stacks or weighted undirected graphs.

If we want to generate valid instances randomly, it is necessary to take these validity constraints into account
during the modelling stage. This is because it would be vanishingly unlikely that an instance randomly sampled from the
unconstrained instance space would be valid. For example, consider the \emph{floor-tile} domain from
Figure~\ref{fig:pddl_domain} and focus on the predicates \texttt{up} and \texttt{down}, which state if a tile
is on top (or bottom) of another. If the \texttt{up} predicate states that a tile is on top of another, the
\texttt{down} predicate should be coherent and correspond with the inverse assignment. If we generate instances
randomly, only a small fraction would satisfy this property for any given pair of tiles, and this fraction shrinks exponentially as the instance size grows.
Another example involves the \texttt{robot-at} predicate, which specifies the tile where a robot is located. It would be
unlikely for randomly generated instances to respect the fact that a robot can only be in one place at one time. 
Almost surely we would end up with instances where a robot is located at various places simultaneously. 
Combining all the implied constraints used by a modeller, it is clear that the chance of randomly generating valid instances is negligibly small.
It is possible to make this claim mathematically precise but we leave this for future work.

\section{Augmented PDDL for Generation of Planning Instances}
\label{sec:augmented_PDDL}

\begin{figure*}[h!]
    \centering
    \includegraphics[width=\linewidth,right]{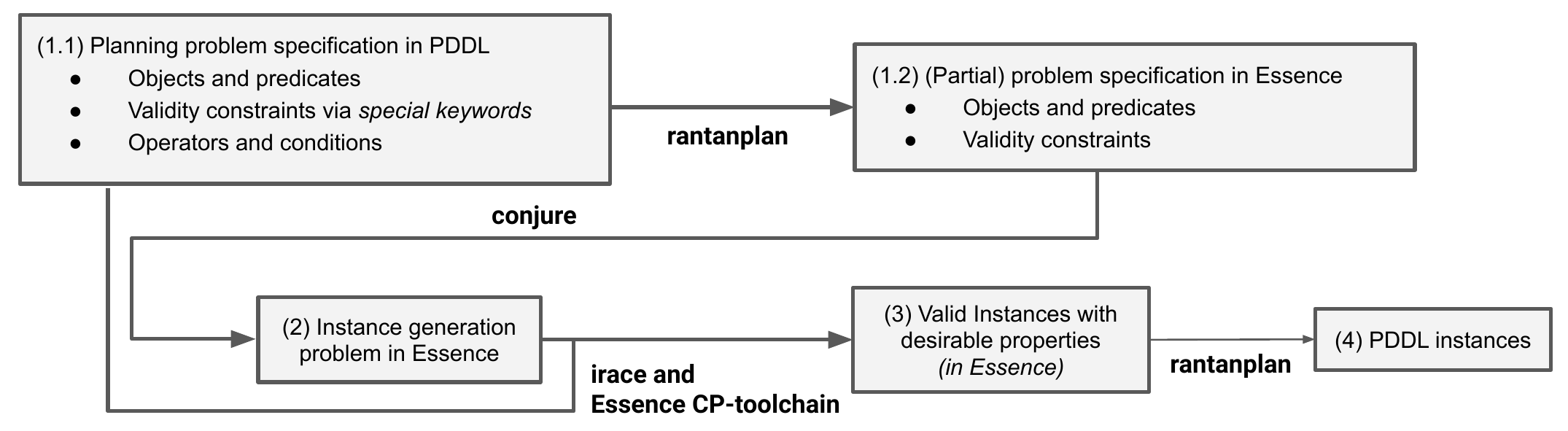}
    \caption{Automated instance generation for planning using Augmented PDDL}
    \label{fig:instGen-PDDL}
\end{figure*}

The \essence language allows expressing validity constraints for a Constraint Programming (CP) problem as \code{where} statements. As shown in Figure~\ref{fig:cp-instGen}, the system we build upon starts by receiving the problem description in \essence with validity constraints as input (step 1), and \conjure will automatically create an instance generator in \essence (step 2), which will then be tuned by \irace in combination with the CP solver \minion to generate instances with desirable properties (step 3). In order to employ the same automated methodology for planning problems, we need to be able to create an instance generator automatically from a planning problem description. Ideally  we would want to use the planning modelling language PDDL to express those validity constraints inside the problem description in step 1 of Figure~\ref{fig:cp-instGen}.

Unfortunately, due to the low-level nature of standard PDDL, the task becomes extremely tedious and error-prone for a human. Firstly, as classical planning only deals with Boolean variables, many-valued variables cannot be directly expressed. This type of variables is essential for modelling common validity constraints such as that a robot can only start at exactly one place.
Secondly, many structural constraints (such as a graph being connected) cannot be expressed in a purely first-order language like PDDL~\cite[Corollary 3.19]{Libkin2004:elements}.
Verifying that a grid specified by adjacency predicates is well-formed, as in our example model, appears also to be impossible to express in pure PDDL, with even restricted versions of this problem still under active investigation~\cite{Kopczynski2019:axiomatizing}.

In this section, we will therefore discuss a solution approach (Figure~\ref{fig:instGen-PDDL}) where we augment PDDL with a new declarative section and extra keywords to allow the modeller to express validity constraints in a PDDL problem description (step 1.1 in Figure~\ref{fig:instGen-PDDL}). The real modelling of those constraints is then generated automatically by \rantanplan~\cite{rantanplan}, a parser specially developed to translate an augmented PDDL model to the high-level language \essence (step 1.2 in Figure~\ref{fig:instGen-PDDL}).
\essence is more expressive than PDDL and allows modelling of structural validity constraints directly.

\subsection{Augmented PDDL for single-predicate validity constraints}
\label{subsec:augmented_pddl_singe_predicate}

\begin{figure}
\centering
\begin{lstlisting}[style=PDDL,breaklines]
  (:instance-constraints
  (init (forall (?r - robot)
    (and  (exactly-k (robot-at ?r _) 1 True) ; a robot starts in a tile 
          (exactly-k (robot-has ?r _) 1 True)))) ; and has one colour
      		 
  ;nothing starts painted
  (init (forall (?t - tile) (exactly-k (painted ?t _) 0 True)))

  ; we are not interested in the clear predicates in the goal state
  (goal (forall (?t - tile) (not (appear (clear ?t)))))
    
\end{lstlisting}
\caption{Snippets of the \code{instance-constraints} section from the \code{floor-tile} domain.}
\label{fig:instance_constraints}
\end{figure}

Validity constraints on the initial and goal states are defined in a new section starting with the new keyword \code{instance-constraints}. 
The following new operands are added for single-predicate validity constraints: \code{init}, \code{goal}, \code{xor}, \code{min}, \code{max}, 
\code{exactly-k}, \code{atleast-k}, \code{atmost-k} and \code{appear}.
By default, when considering numeric functions, the range of values generated goes from 0 to \code{INT\_MAX} (a default upper bound for integer variables, which can be specified by users). The modal operators \texttt{min} and 
\texttt{max} accept the name of a function and an integer. This further restricts the range of
possible values generated. Figure~\ref{fig:instance_constraints} shows an example of the \code{instance-constraints} section for \floor using those new keywords. 

\texttt{init} and \texttt{goal} are modal operators that accept a constraint. This constraint will be then applied to the
initial or  goal state, respectively. 
\texttt{xor} implements the xor logical operation, and has been added for convenience. \texttt{exactly-k},
\texttt{atleast-k} and \texttt{atmost-k} are a family of terms that accept a schematic fluent, a number and a value. 
As their name imply, they restrict the values taken by the subset of grounded variables generated by that schematic fluent.
When specifying these terms, typically we will be interested in one or two parameters of the constrained schematic fluent.
For conciseness, the underscore character (\_) can be used to  indentify parameters that we do not care, acting as a
placeholder in a pattern matching style. Finally, the \texttt{appear} predicate is used to describe the goal state. The initial state is always a total assignment, as per the closed world assumption, but the goal can be a partial assignment. \texttt{appear} can be combined with the \texttt{not} operator to force a state variable
to not appear in the goal state. It is useful to avoid considering non-interesting goals to the modeller.

Not all the new operands are translated directly to \essence. \texttt{min} and \texttt{max} determine the size
of the integer domains of the related PDDL fluents, while \texttt{init} and \texttt{goal} control what is constrained.
The cardinality constraints follow a pattern \texttt{k \{<,>,=\} sum([toInt(x = value) | fluent ])}, where
\texttt{value} is what we are searching for, and the \texttt{fluent} placeholder iterates over the tuples belonging to the
fluent, which is represented as a function. Following the \floor domain, Figure~\ref{fig:example_translation} shows
how a constraint expressing that a given robot can only be at one place in the initial state is translated to \essence.
\begin{figure*}[h!]
    \centering
    \begin{subfigure}{0.45\textwidth}
        \begin{lstlisting}[style=PDDL]
; a robot starts in one tile
(:instance-constraints
(init (forall (?r - robot)
  (exactly-k 
    (robot-at ?r _) 1 True))))
        \end{lstlisting}
        \caption{Constraint expressed in PDDL}
        \label{fig:translation_example_PDDL}
    \end{subfigure}
    \hspace{\fill}
    \begin{subfigure}{0.45\textwidth}
            \begin{lstlisting}[style=essence]
; a robot starts in one tile
forAll var_r : robot .
  1 = sum([toInt(value = true)
    | ((p0,_),value) <-
      init[robot_at],var_r = p0 ]
        \end{lstlisting}
        \caption{Constraint translated to \essence}
        \label{fig:translation_example_essence}
    \end{subfigure}
    \caption{The translation of \code{exactly-k} constraint from PDDL to \essence}
    \label{fig:example_translation}
\end{figure*}

\subsection{Augmented PDDL for structural constraints}
\label{subsec:augmented_pddl_structural}

\begin{figure}[!h]
\centering
\begin{lstlisting}[style=PDDL,breaklines]
(:instance-constraints
  init( isLRUDSquareGrid(tile, up, down, left, right) )
  ... ))
\end{lstlisting} 
\begin{lstlisting}[style=essence, breaklines]
$ ----- Objects and Domains --------
given n_tile: int(1..) 
letting tile be domain int(1 .. n_tile)
$ ---- Auxiliaries ------
given tile_size: int(1..)
where n_tile = tile_size * tile_size
$ ----- Initial State --------
given init: record {
  up : function (total) (tile,tile) --> bool,  
  down : function (total) (tile,tile) --> bool,
  right : function (total) (tile,tile) --> bool,
  left : function (total) (tile,tile) --> bool }
where
  forAll u,v : tile .
    init[up]((u,v))    <-> u = v + tile_size /\
    init[down]((u,v))  <-> u = v - tile_size /\
    init[left]((u,v))  <-> (u = v + 1) /\
      ((u %
    init[right]((u,v)) <-> (u = v - 1) /\ 
      ((u %
\end{lstlisting}
\caption{Expression for the \code{isLRUDSquareGrid} keyword in \essence}
\label{fig:floor-tile-grid-low-level}
\end{figure}

As described in Section~\ref{sec:valid_planning_problems}, another group of validity constraints involves implicit requirements on structures of the underlying map in a planning problem, such as the connections between tiles in the \texttt{floor-tile} problem.
It is possible to express grids quite simply if the relations used to express the grid structure make use of the geometry of the plane.  However, automated instance generation should not restrict the choices made in modelling problems.
In the IPC-14 benchmark dataset, all instances of \texttt{floor-tile} have the tiles forming a square-grid structure and tiles' connections are represented using adjacency relations \texttt{left, right, up} and \texttt{down}. Validity constraints to ensure that these adjacency relations express a square-grid map cannot be efficiently modelled using PDDL due to the limited expressiveness of first-order logic.
Checking that the adjacency relations form a valid grid requires reconstructing geometric information about placement of tiles on the plane. Although it is possible to express the property that the adjacency relations form a grid for special cases, even this requires solving a  challenging tiling problem, and the general case is currently open~\cite{Kopczynski2019:axiomatizing}.

We introduce new PDDL keywords to express those structural constraints, and provide automated translation of those keywords to a CP model in \essence through \rantanplan. An example on expressing a square grid using \{\texttt{left, right, up, down}\} predicates is shown in Figure~\ref{fig:floor-tile-grid-low-level}. 
The newly introduced PDDL keyword \texttt{isLRUDquareGrid} is used and an auxiliary variable indicating the size of the square grid (variable \texttt{tile\_size} in the \essence specification) is generated and will be tuned by \irace during the instance generation process. However, there are two limitations of this automated approach, as we will explain below.

The first limitation is on the \emph{flexibility} to express structural validity constraints. Consider the square-grid structure as an example. There are various ways to define the local connections of cells in a square grid. The choice is normally made based on the specification of planning actions for a specific planning problem. For \texttt{floor-tile}, the locality relations \{\texttt{up, down, left, right}\} are used as the moving actions of a robot at each tile can only follow those directions. However, for the \texttt{city-car} problem, a car can either move horizontally or diagonally. The underlying square-grid map in \texttt{city-car} is then represented using two predicates: \texttt{same\_line} and \texttt{diagonal} for every ordered pair of horizontally and diagonally adjacent cells, respectively. 

The second limitation is about \emph{scalability}. As PDDL predicates are basically boolean functions, the size of validity constraints expressed directly using those predicates can grow very quickly. For example, the square-grid structure expressed in Figure~\ref{fig:floor-tile-grid-low-level} needs to use a nested for loop \code{for u,v: tile}. For a grid size of $20 \times 20$, the total number of constraints is $4 \times 20^{4}$. On a computer with Intel Xeon E5-2640 2.4Ghz CPUs, generating a single instance with such grid size using the CP solver \minion takes about 20 minutes. The time increase also grows fast, as it takes more than 30 minutes to generate a $22 \times 22$ grid.

\section{Expressing the Generation of Planning Problems in \essence}
\label{sec:instance-gen-in-essence}

\begin{figure*}
    \centering
    \includegraphics[width=\linewidth,right]{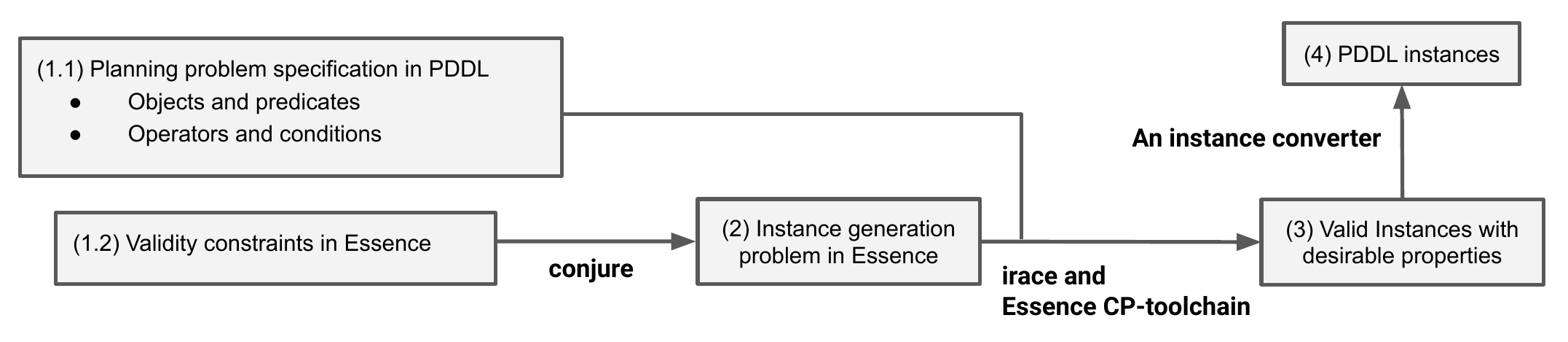}
    \caption{Automated instance generation for planning problems using a hybrid approach, with problem specification in PDDL and validity constraints in \essence. }
    \label{fig:instGen-hybrid}
\end{figure*}

In this section, we discuss an alternative approach to the augmented-PDDL instance generation proposed in the previous section. This is a hybrid approach where users model validity constraints directly in \essence. A summary of this approach is shown in Figure~\ref{fig:instGen-hybrid}. Compared to the previous approach in Figure~\ref{fig:instGen-PDDL}, step 1.2 is now done manually by users. It offers solutions to the two limitations of the augmented-PDDL approach.

Firstly, the new approach allows users to make full use of the high-level modelling language \essence to freely express any validity constraints using constraint modelling, instead of relying on a predefined set of keywords supported by \rantanplan. This overcomes the first limitation on flexibility.

\begin{figure*}
    \includegraphics[width=.8\linewidth]{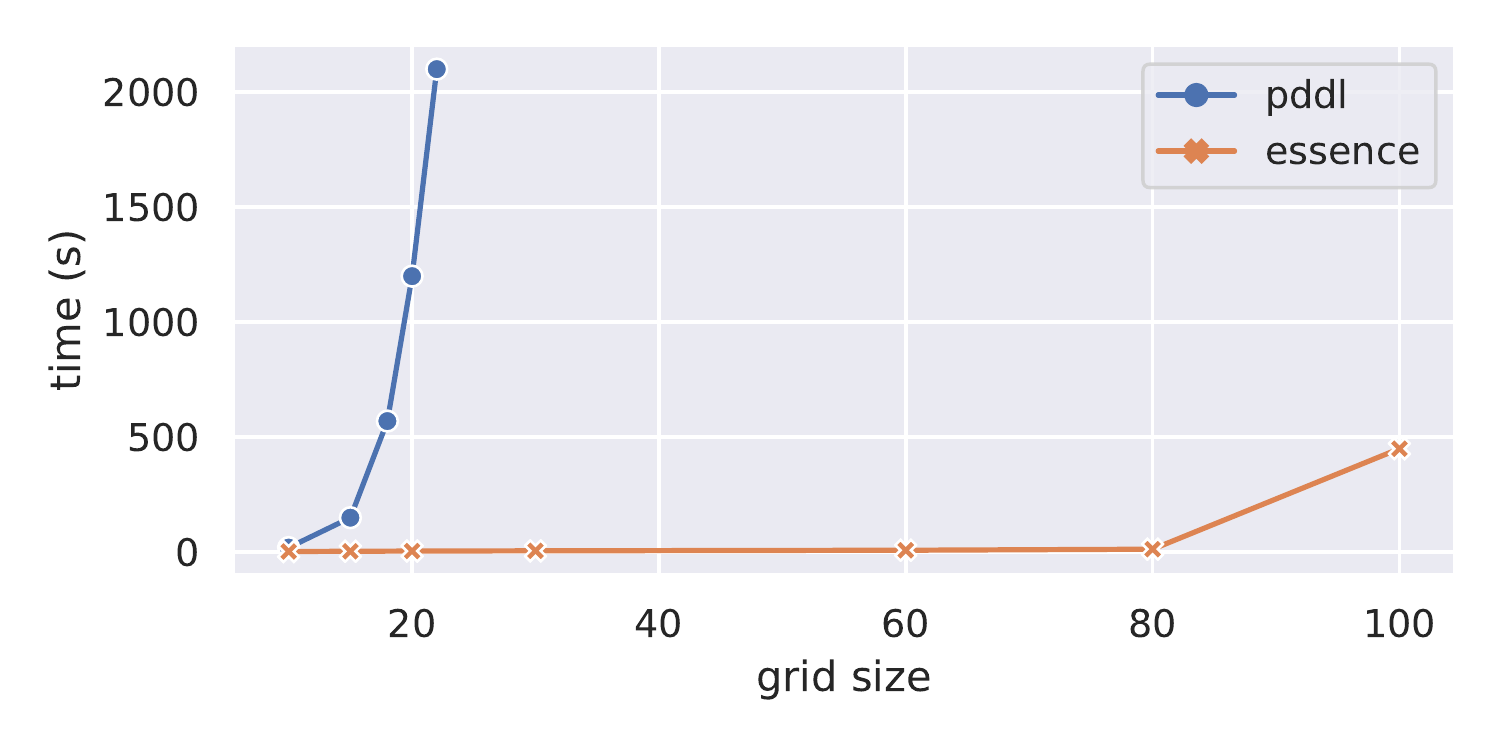}
    \caption{Time to generate floor-tile square-grid instances using our automated instance generation system. The blue line is the approach described in Section~\ref{sec:augmented_PDDL}, and the orange line is the one described in Section~\ref{sec:instance-gen-in-essence}.}
    \label{fig:time}
    \centering
\end{figure*}

Secondly, by modelling validity constraints directly in \essence, users are no longer tied to the relatively low-level boolean representations of PDDL. It is well-known that recovering structure from low-level representation is a difficult and expensive process, as illustrated in the work of Helmert~\cite{sashelmert} where many-valued variables were detected from PDDL representations with Boolean variables. By expressing the constraints directly in \essence, knowledge about implicit structures present in the problem can be easily expressed, which results in more \emph{efficient} representations. For example, in the square-grid structure of \texttt{floor-tile}, 
each of the \texttt{up, down, left, right} relations can be expressed as a function mapping from each tile in the grid to at most one other tile. 
Compared to the low-level representations using boolean functions as in Figure~\ref{fig:instGen-PDDL}, this new representation significantly reduce the number of validity constraints 
(from $4 \times n^4$ to $4 \times n^2$ for any $n \times n$ square grid).
Figure~\ref{fig:time} shows a comparison of the time required by our system to generate square-grid structures for the \texttt{floor-tile} problem using the augmented-PDDL approach and the new one described in this section. The results clearly indicate that the high-level representation (the orange line) significantly improve the efficiency of the instance generation process. 

Another advantage of using high-level representations is that some validity constraints are automatically encoded inside the abstract \essence types themselves without the need of any explicit constraints to express them. An example of the full description of all validity constraints for the \texttt{floor-tile} problem is illustrated in Figure~\ref{fig:floor-tile-instancegen-high-level}. As we can see, the first two validity constraints, which require that each robot can be in only one tile and has one colour at a time, are automatically satisfied thanks to their representations as total functions from \code{robot} to \code{tile} (\code{robot\_at}) and from \code{robot} to \code{color} (\code{robot\_has}).

Despite all the advantages explained above, this approach has a major limitation in terms of \emph{automation}. Instead of writing only one specification for each planning problem of interest, users now have to provide two extra separated inputs to the system. 
The first one is an \essence specification expressing the validity constraints (step 1.2 in Figure~\ref{fig:instGen-hybrid}), with variable names matched with the predicates in the original PDDL problem description (step 1.1 in Figure~\ref{fig:instGen-hybrid}). 
As there are several possible PDDL representations from a high-level abstract description of a problem, the second input is a manually written program to convert the instances in \essence back to the PDDL representation specified in Step 1.1 (from step 3 to step 4 in Figure~\ref{fig:instGen-hybrid}). It is extremely difficult to automate the generation of such converter as the system has to recognise which PDDL representation is the right match.
In the next section, we propose an elegant approach that can overcome all limitations on flexibility and efficiency discussed so far without any trade-offs on automation. 
As we will explain, the whole instance generation process can be fully automated and various encodings, including PDDL, could be supported in a completely transparent manner.

\begin{figure*}[!h]
\centering
\begin{lstlisting}[style=essence, breaklines]
given n_robot : int(1..)
given tile_size: int(1..)
letting n_tile be tile_size*tile_size
given n_color : int(1..)
letting robot be domain int(1 .. n_robot)
letting tile be domain int(1.n_tile)
letting color be domain int(1..n_color)

$ ----- Initial State --------
given init: record {
  robot_at : function (total) robot --> tile,
  robot_has : function (total) robot --> color,
  up : function tile --> tile,   down : function tile --> tile,      
  left : function tile --> tile, right : function tile --> tile,    
  clear : set of tile, available_color : set of color}
where
  forAll c: color . c in init[available_color], $ all colors are available
  forAll u : tile . 
    u in init[clear], $ all titles are clear
    $ square-grid constraints
    u in defined(init[up]) <-> init[up](u) = u - tile_size,
    u in defined(init[down]) <-> init[down](u) = u + tile_size,
    u in defined(init[left]) <-> (init[left](u) = u - 1) 
        /\ (u %
    u in defined(init[right]) <-> init[right](u) = u + 1
        /\ (u %
given goal: record {painted : function (minSize 1) tile --> color}
\end{lstlisting}
\caption{Validity constraints for the whole \texttt{floor-tile} problem expressed directly in \essence}
\label{fig:floor-tile-instancegen-high-level}
\end{figure*}

\section{Abstract Specification of Planning Problems}
\label{sec:essence-for-planning}

This section discusses abandoning a given PDDL description of a planning domain as the starting point for generating instances, and instead extending the \essence language to support the abstract specification of planning problems.

\begin{figure*}
    \centering
    \includegraphics[width=\linewidth,right]{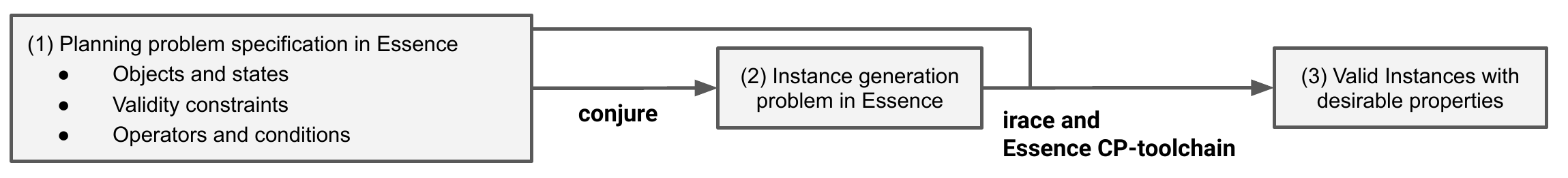}
    \caption{Automated instance generation for planning problems\comment{using the high-level modelling language } in \essence.}
    \label{fig:instGen-essence}
\end{figure*}

The key feature of the \essence language is the provision of high-level type constructors, such as \code{set}, \code{relation} and \code{function}, which allow a problem to be specified directly in terms of the combinatorial structure to be found. Specifying a planning problem in \essence would remove the considerable difficulty of trying to recover this structure from a PDDL description as discussed in the preceding sections. This would simplify the process of instance generation for a planning problem. By providing a refinement of a planning problem specification to a PDDL model, we can also automate much of the work of producing a PDDL encoding of a problem and ensure that the instances generated are synchronised with the model chosen. These advantages result in a straightforward adaptation of the CP automated instance generation system, as presented in Figure~\ref{fig:instGen-essence}.

A simple approach to enabling the specification of planning problems in \essence is to introduce a plan type constructor. In much the same way as PDDL, this would need to support a representation of the objects in the domain, initial and goal states, and plan operators. However, we would gain the far more expressive types available in \essence in order to specify each of these elements. PDDL adopts an implicit frame condition in which all parts of the state not explicitly referenced by an action are assumed to be unchanged, which it also seems sensible to employ in an \essence plan type constructor.

We will illustrate with a hypothetical \essence specification of the \floor problem. Figure \ref{fig:essencePlanSpec} presents the specification of the abstract plan decision variable \code{p} via a new \code{plan} type constructor, which expects four arguments. The first is the state of the planning domain, which is specified using the existing \code{record} type, here capturing the position and colour of each of the robots as well as the current grid state. The initial state is given as a parameter of the same type. The goal, which may only concern a part of the problem state, is flexibly expressed as a set of constraints. For the \code{floor-tile} domain, the goal is a particular grid configuration, expressed as an equality on the grid part of the plan state.

\begin{figure}[t!]
\centering
\begin{lstlisting}[style=essence, breaklines]
given n_robot  : int(1..)
given n_colour : int(1..)
given tile_size : int(1..)
letting CLEAR be 0
letting GRID be domain matrix indexed by [int(1..tile_size), int(1..tile_size)] of int(CLEAR..)
letting COLOUR be new type of size n_colour
letting STATE be domain record { 
  robots : sequence (size n_robot) of record{row : int, column: int, 
                                             colour: COLOUR}, 
  grid : GRID}
given init : STATE
where $ all tiles are clear at the initial state
    forAll i, j : int(1..tile_size) . init[grid][i,j] = CLEAR
given goal : GRID

find p : plan with state STATE
              with initialState init
              with goalState state[grid] = goal
              with actions [goUp, goDown, goLeft, goRight, paintUp, paintDown, changeColour]
\end{lstlisting}
\caption{\label{fig:essencePlanSpec}A possible \essence specification of the \floor problem.}
\end{figure}

The final argument is the list of available actions. Figure \ref{fig:essencePlanAction} presents a hypothetical action representing movement upwards on the grid. The {\tt goUp} action is parameterised on the element of the state that must be selected by the planner, in this case which of the robots to move. Preconditions and effects are expressed as constraints on the parts of the state affected, hence benefiting from the abstract types and operators in \essence. As in PDDL, there is an implied frame condition that any parts of the state not mentioned are unchanged.

\begin{figure}[t!]
\centering
\begin{lstlisting}[style=essence, breaklines]
letting goUp(r in robots) be domain
    action { precondition: grid[r[row]-1, r[column]] = CLEAR,
             effects: r[row]' = r[row]-1}
\end{lstlisting}
\caption{\label{fig:essencePlanAction}A hypothetical \essence plan action compatible with the plan state defined in Fig.~\ref{fig:essencePlanSpec}. The parameter \code{r} is to be chosen by the planner from among those robots in the plan state. The operator \code{'} denotes the state at the subsequent step in the plan. An action such as this could be further annotated with a cost value, if required.}
\end{figure}

An \essence specification of a planning problem such as the one described in this section could be used for both instance generation and automated modelling. The parameters of the plan domain are apparent in the specification of Figure~\ref{fig:essencePlanSpec} and, via the types available in \essence{}, their structure is apparent rather than having to be recovered from a lower level description. 
Most of the validity constraints for the \floor problem are implicitly implied in the high-level representations.
This simplifies instance generation considerably, and would allow a similar approach to that used for \essence specifications of constraint satisfaction/optimisation problems \cite{Akgun2019:instance}.

Automated modelling could also follow the current practice of refining \essence specifications into constraint models using \conjure{}.
With PDDL as the target language, refinement rules would have to be written to encode the \essence types and operators describing the plan state and actions. As is the case for constraint models, multiple refinement rules could be written for the same type to enable alternative PDDL encodings to be generated automatically. A further opportunity would be to exploit the existing \conjure infrastructure to refine the specification to a constraint model of the planning problem, providing alternative solution options via CP, SAT, or SMT through \savilerow.

\section{Conclusion and Future Work}

We have discussed various approaches for adapting a CP automated instance generation system for planning where the problem descriptions are written in PDDL, the standard modelling language for planning problems. The limitations of those approaches are explained and a new language for describing planning problems using \essence, a high-level constraint modelling language, is proposed. Automated instance generation for planning based on \essence offers greater flexibility, efficiency and expressivity compared with its PDDL-based counterpart. In future work, an implementation of the proposal will be provided and a thorough evaluation of such an instance generation system will be done.

\paragraph{Acknowledgements} This work is supported by EPSRC grant EP/P015638/1. Nguyen Dang is a Leverhulme Early Career Fellow. 

\bibliographystyle{splncs04}
\bibliography{main}

\end{document}